\documentclass[conference]{IEEEtran}
\usepackage{cite}
\usepackage{amsmath,amssymb,amsfonts}
\usepackage{algorithmic}
\usepackage{graphicx}
\usepackage{textcomp}
\usepackage{multirow}
\usepackage{array}
\usepackage{url}
\usepackage{xcolor}
\usepackage[hidelinks]{hyperref}
\begin{document}

\title{CALM: Current Aligned Link Manipulation for Single Arm Oversized Object Lifting}

\author{
\IEEEauthorblockN{
Jun Hu\textsuperscript{1,2},
Sihan Chen\textsuperscript{3},
Kosta Jovanovi\'c\textsuperscript{4},
David Navarro-Alarcon\textsuperscript{5},\\
Xueqian Wang\textsuperscript{2},
Jia Pan\textsuperscript{3},
and Peng Zhou\textsuperscript{1,*}
}
\IEEEauthorblockA{
\textsuperscript{1}School of Advanced Engineering, Great Bay University, Dongguan, Guangdong, China\\
\textsuperscript{2}Tsinghua Shenzhen International Graduate School, Tsinghua University, Shenzhen, China\\
\textsuperscript{3}School of Computing and Data Science, The University of Hong Kong, Hong Kong SAR, China\\
\textsuperscript{4}School of Electrical Engineering, University of Belgrade, Belgrade, Serbia\\
\textsuperscript{5}Department of Mechanical Engineering, The Hong Kong Polytechnic University,\\
Kowloon, Hong Kong SAR, China
}
}

\maketitle

\begin{abstract}

Most robots manipulate objects solely with their end effectors, whereas humans flexibly leverage different body parts, such as the forearm and elbow, especially when handling oversized objects. Learning such whole-arm manipulation is challenging due to long-horizon sparse rewards, limited contact sensing, and the sim-to-real gap in contact and actuator dynamics. To address these challenges, we propose Current-Aligned Link Manipulation, a framework for learning long-horizon contact-rich manipulation using motor current as joint load related feedback. Three stage-specific policies first learn repositioning, grasping, and lifting using privileged simulation information, and a stage router sequences them to generate complete task demonstrations. For sim-to-real transfer, a causal current mapper predicts physical motor current from simulated joint histories, aligning the actuator current observation between simulation and hardware. A unified student policy then learns from these demonstrations using only deployable sensor observations and is further refined with DAgger. The task policies are trained entirely in simulation, and the final student is deployed on hardware. Experiments demonstrate 76.2\% (762/1000 trials) complete-task success in simulation and 73.3\% success (22/30 trials) on the physical robot for sequential oversized-object lifting.

\end{abstract}

\begin{IEEEkeywords}
simulation to reality transfer, whole arm manipulation, oversized object lifting, reinforcement learning, imitation learning, industrial robots
\end{IEEEkeywords}

\section{Introduction}
\label{sec:introduction}
\begin{figure}[t]
    \centering
    \includegraphics[width=\columnwidth]{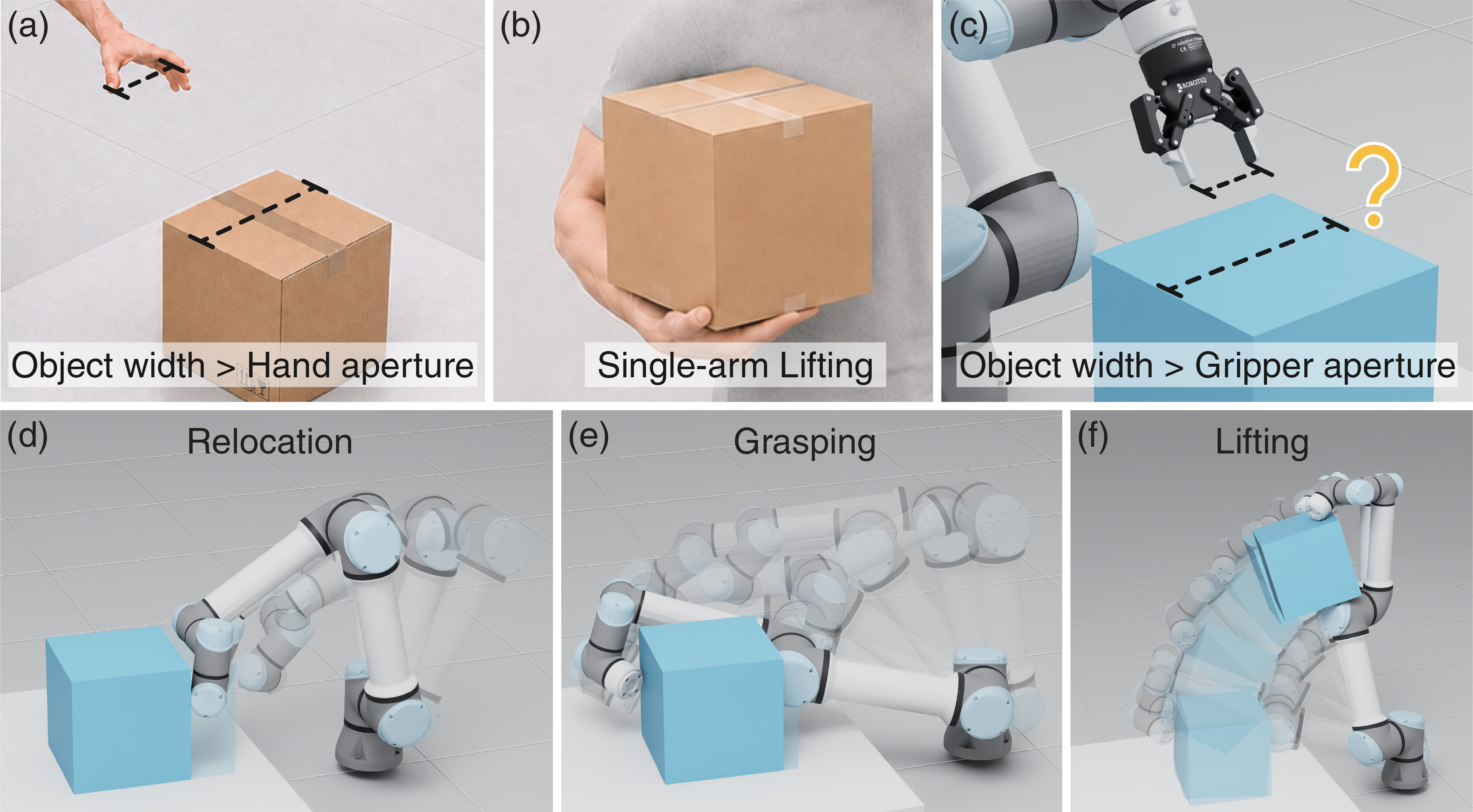}
    \caption{Motivation and task overview. (a) When an object's width exceeds the functional grasp aperture of one hand, (b) a person can recruit the whole arm to support and carry it. Analogously, (c) when the object's width exceeds the aperture of the robot's end effector gripper, we investigate whether a single rigid manipulator can use intermediate arm links to (d) reposition the object, (e) establish an opposed grasp, and (f) lift it. }
    \label{fig:motivation}
\end{figure}

Humans routinely exploit different parts of the arm to interact with objects, particularly when an object cannot be grasped by one hand. In contrast, robotic manipulation predominantly relies on end-effector grippers, while oversized or heavy objects often require specialized grippers or coordinated dual-arm systems \cite{glick2018,zhou2026}. This reliance limits the manipulation capability of conventional robot arms, whose intermediate links are typically treated as collision geometry rather than potential contact surfaces. In this work, we investigate whether a standard rigid manipulator can exploit its intermediate links to reposition, grasp, and lift an oversized object using a single arm, without adding tactile or force/torque sensors to the robot, as illustrated in \hyperref[fig:motivation]{Fig.~\ref*{fig:motivation}}.

Lifting an oversized box with a standard single  robot arm 
is a long horizon, sequential, and contact rich task. Repositioning, grasping, and lifting must occur in order, and each stage depends on contact feedback. This setting presents two research problems. First, tactile sensors can provide contact information \cite{zhou2026vision}, but distributed whole arm tactile sensing usually requires specialized sensorized hardware \cite{punyo2022,barreiros2025}. Motor current offers an alternative source of joint load related feedback \cite{wahrburg2018}. However, simulated joint effort does not directly match physical motor current because friction, drivetrain losses, backlash, and low level control are not modeled exactly \cite{tan2018,hwangbo2019}. This mismatch makes motor current difficult to use consistently for policy learning in simulation and execution on hardware. Second, imitation learning and reinforcement learning both face limitations in this task. Imitation learning depends on demonstrations \cite{argall2009}, but visual teleoperation does not provide the tactile feedback needed to regulate whole arm contact \cite{yu2025mimictouch}. Direct reinforcement learning on hardware requires extensive interaction, repeated resets, and potentially unsafe exploration \cite{kober2013}. Training in simulation avoids these physical costs, but model and actuator errors can prevent the learned policy from transferring to the real robot \cite{hwangbo2019}.

\begin{figure*}[t]
    \centering
    \includegraphics[width=0.90\textwidth]{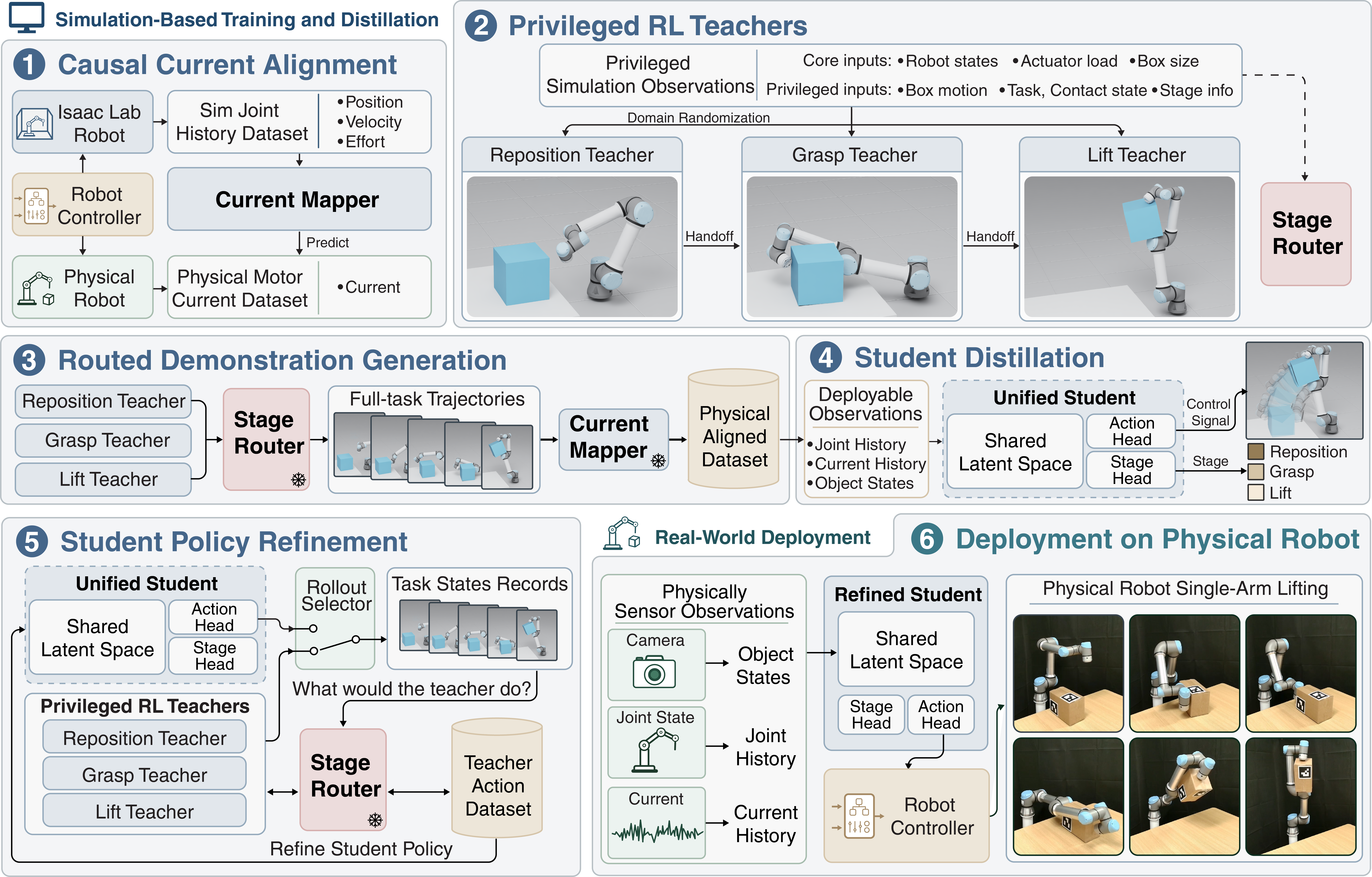}
    \caption{Overview of CALM. A causal Transformer maps simulated joint histories to physical motor current, while three privileged teachers learn repositioning, grasping, and lifting. During complete task rollouts, privileged simulation state selects the active teacher to produce complete trajectories under domain randomization. These trajectories supervise one stage aware student, which is refined with DAgger and deployed with deployable observations only.}
    \label{fig:pipeline}
\end{figure*}

To address these challenges, we propose the Current Aligned Link Manipulation (CALM) framework, as shown in \hyperref[fig:pipeline]{Fig.~\ref*{fig:pipeline}}. CALM uses a causal current mapper to predict physical motor current from simulated command, joint-state, and joint-effort histories. Using these aligned current observations, CALM distills stage-specific privileged teachers into a unified student policy for deployment on the physical robot. For the present task, the teachers learn repositioning, grasping, and lifting in simulation and generate complete demonstrations under domain randomization. These demonstrations and subsequent DAgger refinement train a unified student policy that performs the physical task using only physically available sensor observations.

This work makes the following contributions:
\begin{enumerate}
\item A single-arm link-manipulation system that intentionally uses rigid intermediate robot links to form an opposed grasp and lift objects exceeding the end-effector grasp aperture, without specialised grasping hardware or distributed whole-arm tactile sensing;
\item a causal current alignment method that maps simulated joint histories to the motor current observations expected on the physical robot, enabling target task policy learning entirely in simulation while reducing the proprioceptive gap between simulation and reality; and
\item a stage routed teacher to student framework that enables independent, stage specific training and fine tuning of teacher policies. It connects the teachers through feasible handoff states to generate complete task demonstrations, which are distilled into a unified deployable student using stage supervision and DAgger.
\end{enumerate}

\section{Related Work}
\label{sec:related_work}

\subsection{Whole Arm and Robot Link Manipulation}

Whole arm manipulation uses intermediate robot links as contact surfaces instead of treating them only as collision geometry. Classical work analyzes force decomposition for enveloping grasps with contacts distributed across limbs \cite{bicchi1994}. Compliant humanoid control later demonstrated object manipulation using contacts on the arms and chest \cite{florek2014}. Other systems use customized curved link surfaces to immobilize objects \cite{seo2016}, while continuum manipulators exploit compliant backbones that conform around the object \cite{walker2013}. These studies establish the mechanics and utility of link contact, but commonly use multiple limbs, or rely on morphology designed for conformal grasping.

Learning based systems extend whole arm manipulation to uncertain and contact rich tasks. Yuan et al. use reinforcement learning and a topology based representation to hold and transport people or bulky objects with robot arms and torso geometry \cite{yuan2019}. Punyo-1 combines two compliant tactile arms, soft end effectors, and a sensorized chest for large object manipulation \cite{punyo2022}. Barreiros et al. subsequently use example guided reinforcement learning and domain randomization to acquire large object skills on this compliant upper body platform with pressure sensing \cite{barreiros2025}. Other recent systems exploit quadrupedal whole body contact, distributed whole body vision, or learned actuator models for athletic loco manipulation \cite{jeon2024,robopanoptes2025,uan2025}. Overall, existing systems predominantly exploit multiple limbs or custom compliant and sensing hardware and generally emphasize short horizon behaviors. CALM addresses the resulting gap by using one rigid arm with six axes to reposition, grasp, and lift an oversized object through link contact.

\subsection{Current Alignment for Sim2real Transfer}

Distributed tactile sensing is not commonly available on conventional industrial manipulators, whereas joint motor current is broadly available through robot controllers. Motor current therefore provides an accessible joint load related feedback signal. Model based methods relate motor current, joint state, and robot dynamics to external joint torques and Cartesian loads \cite{wahrburg2018,mohammad2023}. However, unmodeled robot dynamics make the relationship between current and joint load difficult to represent accurately. Temporal learning has been proposed to reduce dependence on precise dynamics models \cite{callar2023}, and learned actuator models capture unmodeled actuation effects for control transferred from simulation to reality \cite{tan2018,hwangbo2019}. Existing methods primarily address load representation or actuator modeling rather than observation alignment; CALM instead causally aligns simulated joint histories with physical motor current observations, providing a consistent current representation in simulation and on hardware to reduce the policy's proprioceptive gap between simulation and reality.

\subsection{Stage Routed Policy Transfer for Long Horizon Tasks}

Long horizon tasks are commonly decomposed into stage specific policies. Sequential Dexterity chains dexterous skills through learned transition feasibility \cite{chen2023}. Robot Parkour uses DAgger to distill privileged locomotion teachers into one vision based policy \cite{zhuang2023}, and RLDG distills task specific reinforcement learning experts into a generalist manipulation policy \cite{xu2025}. DAgger reduces imitation covariate shift by querying teachers at states visited by the student \cite{dagger2011}. These methods demonstrate skill chaining and multi teacher distillation, but primarily address dexterous manipulation, locomotion, or collections of distinct tasks. In a physically coupled contact sequence, each stage changes the feasible initial conditions of the next. Because independently trained teachers may not preserve feasible transitions, CALM uses privileged stage routing to connect their handoff states and distill them into a unified deployable policy.

\section{Method}
\label{sec:method}

We propose CALM to address the task of lifting oversized boxes with a single robot arm. The overall framework is illustrated in \hyperref[fig:pipeline]{Fig.~\ref*{fig:pipeline}}. The framework comprises six components: (1) \emph{Causal Current Alignment} maps simulated histories to predicted physical current observations; (2) \emph{Privileged Teacher Training} trains three teachers using privileged simulation state; (3) \emph{Stage Conditioned Rollout Collection} routes the teachers to generate complete trajectories; (4) \emph{Student Distillation} trains one policy to imitate the routed teacher behaviors using only deployable observations; (5) \emph{DAgger Refinement} adds teacher corrections at student visited states; and (6) \emph{Physical Deployment} deploys the final student policy on the physical robot. 
\subsection{Task and Observation Formulation}
\label{subsec:task_formulation}

We structure the box lifting task as three sequential stages, as illustrated in \hyperref[fig:motivation]{Fig.~\ref*{fig:motivation}(d)--(f)}: (1) \emph{repositioning}, in which the robot moves the box into a configuration suitable for link contact; (2) \emph{grasping}, in which designated regions of UR7e wrist~3 and the upper arm link establish opposed contact with the box; and (3) \emph{lifting}, in which the robot raises the box to the target height. A trial succeeds when the box bottom remains at least $0.20$~m above the table for $5$~s. Other contacts between the robot and box or between the robot and table are treated as unintended.

The three privileged teacher policies are trained with PPO in simulation. A student policy is subsequently trained in simulation to imitate the routed teachers using only sensor observations compatible with deployment. During hardware deployment, measured motor current and a camera estimated box pose replace the corresponding simulation derived channels while retaining the same ordering and normalization. The observations used for teacher training, student training, and hardware deployment are summarized in Table~\ref{tab:observation_contract}.

For the policy equations, $t$ indexes the $30$~Hz control steps, and $\mathcal{J}=\{R,G,L\}$ denotes the repositioning, grasping, and lifting stages. The router stage is $z_t\in\mathcal{J}$, with one hot encoding $\mathbf{e}(z_t)\in\mathbb{R}^{3}$. Joint position and velocity are $\mathbf{q}_t,\dot{\mathbf{q}}_t\in\mathbb{R}^{6}$. For teacher $j\in\mathcal{J}$, $\mathbf{a}^{j}_t\in\mathbb{R}^{6}$ is its command, $\mathbf{a}^{*}_t=\mathbf{a}^{z_t}_t$ is the router selected teacher command, and $\mathbf{a}_t\in\mathbb{R}^{6}$ is the executed command. Superscripts $B$ and $T$ denote the box and task frames; subscripts $w$, $u$, and $b$ denote wrist~3, the upper arm link, and the box. A tilde marks a normalized quantity, a hat marks an estimate or prediction, and superscript $D$ identifies the deployable observation interface.

\begin{table*}
\begingroup
\caption{Observation variables for privileged teacher training, student imitation training, and physical deployment.}
\label{tab:observation_contract}
\centering
\footnotesize
\setlength{\tabcolsep}{4pt}
\begin{tabular}{>{\centering\arraybackslash}p{0.095\textwidth}p{0.30\textwidth}p{0.235\textwidth}p{0.265\textwidth}}
\hline
\multicolumn{1}{c}{\parbox[c][0.9cm][c]{0.095\textwidth}{\centering\bfseries Observation}} &
\multicolumn{1}{c}{\parbox[c][0.9cm][c]{0.30\textwidth}{\centering\bfseries Teacher PPO training\\(simulation)}} &
\multicolumn{1}{c}{\parbox[c][0.9cm][c]{0.235\textwidth}{\centering\bfseries Student imitation training\\(simulation)}} &
\multicolumn{1}{c}{\parbox[c][0.9cm][c]{0.265\textwidth}{\centering\bfseries Hardware deployment}} \\
\hline
\textbf{Core inputs} &
(1) Joint position $\mathbf{q}_t$, joint velocity $\dot{\mathbf{q}}_t$, and previous teacher action $\mathbf{a}^{j}_{t-1}$; \newline
(2) normalized simulated joint effort $\widetilde{\boldsymbol{\tau}}^{\mathrm{sim}}_t$; \newline
(3) normalized simulated box dimensions $\widetilde{\mathbf{s}}_b$
&
(1) Joint position $\mathbf{q}_t$, joint velocity $\dot{\mathbf{q}}_t$, and previous executed action $\mathbf{a}_{t-1}$; \newline
(2) normalized mapped current $\widetilde{\mathbf{I}}_t$; \newline
(3) estimated pose $(\hat{\mathbf{p}}^{T}_t,\hat{\boldsymbol{\rho}}^{T}_t)$ and normalized dimensions $\widetilde{\mathbf{s}}_b$
&
(1) Encoder joint position $\mathbf{q}_t$, joint velocity $\dot{\mathbf{q}}_t$, and previous action $\mathbf{a}_{t-1}$; \newline
(2) normalized measured current $\widetilde{\mathbf{I}}_t$; \newline
(3) camera estimated pose $(\hat{\mathbf{p}}^{T}_t,\hat{\boldsymbol{\rho}}^{T}_t)$ and normalized measured dimensions $\widetilde{\mathbf{s}}_b$ \\
\hline
\textbf{Privileged inputs} &
(1) Link positions relative to the box $(\mathbf{p}^{B}_{w,t},\mathbf{p}^{B}_{u,t})$ and box velocity $(\mathbf{v}^{T}_{b,t},\boldsymbol{\omega}^{T}_{b,t})$; \newline
(2) box clearance $\widetilde{h}_t$, planar target error $\widetilde{\boldsymbol{\varepsilon}}^{xy}_t$, and contact forces $(\widetilde{\mathbf{f}}^{B}_{w,t},\widetilde{\mathbf{f}}^{B}_{u,t})$; \newline
(3) true stage $\mathbf{e}(z_t)$ and teacher specific terms $\boldsymbol{\xi}^{R}_t$, $\boldsymbol{\xi}^{G}_t$, and $\boldsymbol{\xi}^{L}_t$
& -- & -- \\
\hline
\end{tabular}
\endgroup
\end{table*}

The repositioning, grasping, and lifting teachers receive privileged observations $\mathbf{o}^{R}_t\in\mathbb{R}^{53}$, $\mathbf{o}^{G}_t\in\mathbb{R}^{54}$, and $\mathbf{o}^{L}_t\in\mathbb{R}^{57}$, respectively. Each teacher observation combines the core policy inputs and additional privileged inputs summarized in Table~\ref{tab:observation_contract}. The additional privileged inputs include the following teacher specific terms:
\begin{equation}
\begin{aligned}
\boldsymbol{\xi}^{R}_t &={}
[\sin(4\varepsilon^{\psi}_t),\cos(4\varepsilon^{\psi}_t)]\in\mathbb{R}^{2},\\
\boldsymbol{\xi}^{G}_t &={}
[\chi^{G}_t,\widetilde{n}^{G}_{\mathrm{hold},t},
\widetilde{n}^{G}_{\mathrm{loss},t}]\in\mathbb{R}^{3},\\
\boldsymbol{\xi}^{L}_t &={}
[\gamma_t,\kappa_t,\widetilde{n}^{L}_{\mathrm{confirm},t},
\widetilde{n}^{L}_{\mathrm{loss},t},
\widetilde{n}^{L}_{\mathrm{hold},t},
\mathbb{I}[h_t\geq h^{\star}]]\in\mathbb{R}^{6},
\end{aligned}
\label{eq:teacher_stage_observation}
\end{equation}
where $\varepsilon^{\psi}_t$ is the box yaw alignment error; $\chi^{G}_t$ is the grasp latch flag; $\gamma_t$ and $\kappa_t$ are grasp quality and confirmation; the $\widetilde{n}$ terms are normalized confirmation, hold, or loss counters; $h^{\star}$ is the target lift clearance; and $\mathbb{I}[\cdot]$ is the indicator function.

The deployable observation is
\begin{equation}
\mathbf{o}^{D}_t =
\left[\mathbf{q}_t,\;\dot{\mathbf{q}}_t,\;
\widetilde{\mathbf{I}}_t,\;\hat{\mathbf{p}}^{T}_t,\;
\hat{\boldsymbol{\rho}}^{T}_t,\;\widetilde{\mathbf{s}}_b,\;\mathbf{a}_{t-1}\right]
\in\mathbb{R}^{34}.
\label{eq:deployable_observation}
\end{equation}
Here, $\widetilde{\mathbf{I}}_t\in\mathbb{R}^{6}$ is the normalized current channel, and $\hat{\mathbf{p}}^{T}_t\in\mathbb{R}^{3}$ and $\hat{\boldsymbol{\rho}}^{T}_t\in\mathbb{R}^{4}$ are the estimated box position and quaternion. Table~\ref{tab:observation_contract} specifies how the shared channels are instantiated during student training and hardware deployment. The observations are stacked into the causal history
\begin{equation}
\mathbf{O}^{D}_t =
\left[\mathbf{o}^{D}_{t-K+1},\ldots,\mathbf{o}^{D}_t\right]
\in\mathbb{R}^{K\times34},\qquad K=30.
\label{eq:deployable_history}
\end{equation}
The history $\mathbf{O}^{D}_t$ is supplied to the student and provides one second of causal context.

\subsection{Causal Current Alignment}
\label{subsec:current_alignment}

Causal current alignment uses a Transformer \cite{vaswani2017attention}, termed the current mapper in Fig.~\ref{fig:pipeline}, to predict the physical robot motor current trajectory from simulated command, joint state, and joint effort histories. It is designed to reduce an observation gap between simulation and reality: under matched executions, joint positions are comparatively consistent between simulation and the physical robot, whereas simulated joint effort does not reproduce physical motor current because actuator, transmission, friction, and low level control effects are simplified. Motor current serves as a joint load related feedback signal, and CALM aligns this observation channel before student training.

To collect paired training data, we first defined a safe Cartesian workspace and sampled random waypoints within it. MoveIt connected successive waypoints with time parameterized trajectories whose joint velocities varied; identical commanded joint positions and velocities were sent to the simulated and physical robots. We synchronously recorded the commanded joint positions and velocities. The simulation stream comprised joint positions, velocities, and joint efforts, whereas the physical robot stream comprised joint positions, velocities, and motor currents.

A Transformer based current mapper was trained on this synchronized dataset. Let $m$ denote the index of samples recorded at $120$~Hz for current mapping. Its input vector, causal history, and forecast are
\begin{equation}
\begin{aligned}
\mathbf{x}_m={}&[\mathbf{q}^{\mathrm{cmd}}_m,\dot{\mathbf{q}}^{\mathrm{cmd}}_m,
\mathbf{q}^{\mathrm{sim}}_m,\dot{\mathbf{q}}^{\mathrm{sim}}_m,
\boldsymbol{\tau}^{\mathrm{sim}}_m]\in\mathbb{R}^{30},\\
\mathbf{X}_{m-H+1:m}={}&
[\mathbf{x}_{m-H+1},\ldots,\mathbf{x}_m]\in\mathbb{R}^{H\times30},\\
\widehat{\mathbf{I}}^{\mathrm{phys}}_{m+1:m+F}={}&
f_{\phi}(\mathbf{X}_{m-H+1:m})\in\mathbb{R}^{F\times6}.
\end{aligned}
\label{eq:current_mapper_io}
\end{equation}
Here, superscripts $\mathrm{cmd}$, $\mathrm{sim}$, and $\mathrm{phys}$ denote commanded, simulated, and physical signals, respectively; $\boldsymbol{\tau}^{\mathrm{sim}}_m$ is simulated joint effort; and $f_{\phi}$ is the current mapper with parameters $\phi$. The training target $\mathbf{I}^{\mathrm{phys}}_{m+1:m+F}$ is the corresponding measured motor current. The mapper uses $H=120$ samples, corresponding to $1$~s, to predict $F=60$ samples, corresponding to $0.5$~s. During student training, the current mapper is evaluated at $120$~Hz, and the $30$~Hz student receives the latest mapped current after every four mapper updates. A learned linear projection followed by layer normalization and a GELU activation maps each $\mathbf{x}_m$ to an embedding with 128 dimensions. Learned positional embeddings are then added before four causally masked Transformer encoder layers, each with eight attention heads. A forecast head maps the final encoded token to all six current channels over the prediction horizon. The mapper is trained using the multistep current alignment loss
\begin{equation}
\mathcal{L}_{\mathrm{map}}=
\mathcal{L}_{\mathrm{Huber}}
+\lambda_{\Delta}\mathcal{L}_{\Delta}
+\lambda_{\Delta^2}\mathcal{L}_{\Delta^2},
\label{eq:current_loss}
\end{equation}
where $\mathcal{L}_{\mathrm{Huber}}$ compares the predicted and measured current magnitudes, $\mathcal{L}_{\Delta}$ and $\mathcal{L}_{\Delta^2}$ penalize mismatch in their first and second temporal differences, and $\lambda_{\Delta}$ and $\lambda_{\Delta^2}$ are the corresponding weights.

After training, the mapper is frozen. During routed demonstrations and DAgger, its normalized prediction supplies $\widetilde{\mathbf{I}}_t$ in Eq.~\ref{eq:deployable_observation}; during deployment, normalized measured current supplies the same channel. The mapper performs observation domain alignment while leaving the simulator dynamics unchanged.

\subsection{Privileged Teachers and Router}
\label{subsec:teachers_router}

We train three privileged PPO teachers in Isaac Lab: $\pi_R$ repositions and settles the box, $\pi_G$ establishes the opposed grasp, and $\pi_L$ lifts the box. All teachers share the action interface and control rate, and their privileged observations are summarized in Table~\ref{tab:observation_contract}. During PPO training, domain randomization varies the box dimensions and contact friction together with other physics, controller, and initial state parameters. The randomized box dimensions are supplied to each teacher through $\widetilde{\mathbf{s}}_b$, allowing the policy to adapt its behavior to object geometry. Each teacher uses a shaped reward specific to its stage, whose principal positive and penalty terms are summarized in Table~\ref{tab:teacher_rewards}. The teachers are trained sequentially to reproduce the state distribution encountered at each handoff. Successful repositioning terminal states produced by $\pi_R$ serve as reset states for training the grasp teacher $\pi_G$; similarly, successful grasping terminal states produced by $\pi_G$ serve as reset states for training the lift teacher $\pi_L$. Thus, each downstream teacher learns from states that the preceding teacher can produce.

Each teacher uses separate actor and critic MLPs with hidden dimensions $[256,256,128]$ and ELU activations. PPO uses 512 environments and Adam, with a reward discount factor of $0.99$ and a GAE factor of $0.95$. Three teachers train for 50,000 iterations at learning rate $3\times10^{-4}$.

\begin{table}[t]
\caption{Principal rewards and penalties specific to each stage.}
\label{tab:teacher_rewards}
\centering
\footnotesize
\setlength{\tabcolsep}{3pt}
\begin{tabular}{p{0.2\columnwidth}p{0.75\columnwidth}}
\hline
\textbf{Teacher} & \textbf{Reward and penalty terms} \\
\hline
Reposition $\pi_R$ & \textbf{Reward:} (1) target progress; (2) push face approach; (3) aligned contact; (4) settling. \textbf{Penalty:} (1) edge contact or excessive force; (2) action magnitude/rate; (3) shoulder deviation. \\
Grasp $\pi_G$ & \textbf{Reward:} (1) contact at designated links; (2) centered opposed grasp; (3) stable hold. \textbf{Penalty:} (1) missing or invalid contact; (2) box motion; (3) grip loss; (4) table contact; (5) motion and force variation after latching. \\
Lift $\pi_L$ & \textbf{Reward:} (1) grasp confirmation and maintenance; (2) lift progress and height; (3) settling and holding. \textbf{Penalty:} (1) weak or lost grasp; (2) premature or stalled lifting; (3) lateral motion; (4) wrist slip; (5) large actions before confirmation. \\
\hline
\end{tabular}
\end{table}

A deterministic privileged router selects one teacher at each simulation step. It chooses $\pi_R$ when the box is outside the settled grasp region, $\pi_G$ when repositioning is valid but the opposed grasp is not, and $\pi_L$ after a stable grasp is established. Unsafe or unrecoverable states terminate the episode. Persistence, hysteresis, and minimum dwell times prevent rapid switching, while loss of position or grasp returns control to the corresponding earlier teacher.

\subsection{Routed Demonstrations and Domain Randomization}
\label{subsec:demonstrations}

Routed demonstrations provide imitation samples $(\mathbf{O}^{D}_t,\mathbf{a}^{*}_t,z_t)$. The samples are collected from complete repositioning, grasping, and lifting trajectories generated by the three frozen teachers under router control. Only $\mathbf{o}^{D}_t$ is recorded for student input, and the history $\mathbf{O}^{D}_t$ remains continuous across teacher switches. The stage label is a supervision target only and is not included in the deployed observation.

To generate the imitation dataset, complete routed teacher episodes are collected under domain randomization. Predefined randomization ranges are applied to box dimensions, position and orientation, contact friction, delays, and sensor noise. For the final student training dataset, only samples from complete successful episodes will be retained. Each retained episode will preserve the routed sequence of the three stages rather than merge independently collected stage segments.

\subsection{Stage Aware Student and DAgger}
\label{subsec:student}

A per step encoder and GRU with parameters $\theta$ map $\mathbf{O}^{D}_t$ to the shared latent state $\mathbf{h}_t=f_{\theta}(\mathbf{O}^{D}_t)$. The action head predicts $\hat{\mathbf{a}}_t=g_a(\mathbf{h}_t)\in\mathbb{R}^{6}$, while the auxiliary stage head produces logits $\boldsymbol{\ell}_t=g_z(\mathbf{h}_t)\in\mathbb{R}^{3}$. Because both heads operate on $\mathbf{h}_t$, stage supervision encourages the temporal encoder to represent information specific to each stage that also informs action prediction. The stage label is used only during training; neither the label nor the predicted stage selects a teacher or action head at deployment.

The student maps each 34 dimensional observation to 256 features and processes the history with a two layer GRU of width 256. Its action and stage heads have dimensions $[256,256,6]$ and $[256,128,3]$. Training uses AdamW for 40 epochs with batch size 1,024, and learning rate $3\times10^{-4}$. Three DAgger rounds use $\beta=0.5$, $0.3$, and $0.1$.

The behavior cloning objective is
\begin{equation}
\mathcal{L}_{\mathrm{student}}=
\underbrace{w_{z_t}\frac{1}{6}
\left\|\hat{\mathbf{a}}_t-\mathbf{a}^{*}_t\right\|_2^2}_{\mathcal{L}_{\mathrm{act}}}
+\lambda_z\underbrace{\mathrm{CE}(\boldsymbol{\ell}_t,z_t)}_{\mathcal{L}_{\mathrm{stage}}},
\label{eq:student_loss}
\end{equation}
where the factor $1/6$ averages the squared action error across the six joints, $w_{z_t}$ compensates for unequal sample frequencies across stages, and $\lambda_z$ controls the auxiliary stage loss. The action term distills the selected teacher's command, whereas the cross entropy term trains the shared representation to distinguish the three task stages.

After behavior cloning, DAgger refines the student through randomized simulation rollouts. During these rollouts, the student receives $\mathbf{O}^{D}_t$, including predicted physical current, whereas the selected teacher receives $\mathbf{o}^{z_t}_t$, including simulated joint effort and exact task state. At each step, the privileged router selects the teacher, and both that teacher and the student predict an action. A Bernoulli selector advances the simulation with $\mathbf{a}^{*}_t$ with probability $\beta$ and with $\hat{\mathbf{a}}_t$ otherwise. This selection affects only the rollout trajectory. At every visited state, the deployable observation history $\mathbf{O}^{D}_t$, the teacher's predicted action $\mathbf{a}^{*}_t$, and the router's stage label $z_t$ are recorded as a training sample, regardless of which action is executed. The resulting samples are aggregated with the existing dataset to refine the student policy.

\subsection{Physical Deployment}
\label{subsec:deployment}

At deployment, the simulated joint state, predicted current, and simulated box pose are replaced by measured joint state, measured motor current, and a camera estimated box pose in the same task frame. Measured box dimensions are supplied through the same normalized task input channels used in simulation. Channel order, units, normalization, history initialization, observation rate, and action interface are identical in simulation and on hardware. Independent safety limits constrain robot speed, workspace, joint configuration, and motor current during execution. No teacher policy, router, privileged stage, exact simulation state, or simulated contact signal is used.

\section{Experiments and Results}
\label{sec:experiments}

The experiments are designed to answer three research questions: (1) Does current alignment reduce the observation gap between simulation and reality? (2) Do stage aware distillation and DAgger improve unified student performance? (3) How reliably does the complete CALM system perform physical box lifting? These questions and the corresponding experiments are summarized in Table~\ref{tab:question_map}.

\begin{table}[t]
\caption{Research questions and corresponding experiments.}
\label{tab:question_map}
\centering
\footnotesize
\setlength{\tabcolsep}{4pt}
\begin{tabular}{p{0.27\columnwidth}p{0.64\columnwidth}}
\hline
\textbf{Question summary} & \textbf{Experiments providing evidence} \\
\hline
Current alignment & Secs.~\ref{subsec:exp_alignment} and~\ref{subsec:exp_physical}: current prediction accuracy and 10 hardware trials without current alignment \\
Student distillation & Secs.~\ref{subsec:exp_teachers} and~\ref{subsec:exp_student}: controlled comparison of the full student policy and its ablated variants \\
Physical CALM & Sec.~\ref{subsec:exp_physical}: 30 hardware trials of the final CALM policy and a current alignment ablation comprising 10 additional trials, with a new random box placement before each trial \\
\hline
\end{tabular}
\end{table}

\subsection{Experimental Setup and Evaluation Protocol}
\label{subsec:exp_protocol}

\begin{figure}[!b]
    \centering
    \includegraphics[width=0.9\columnwidth]{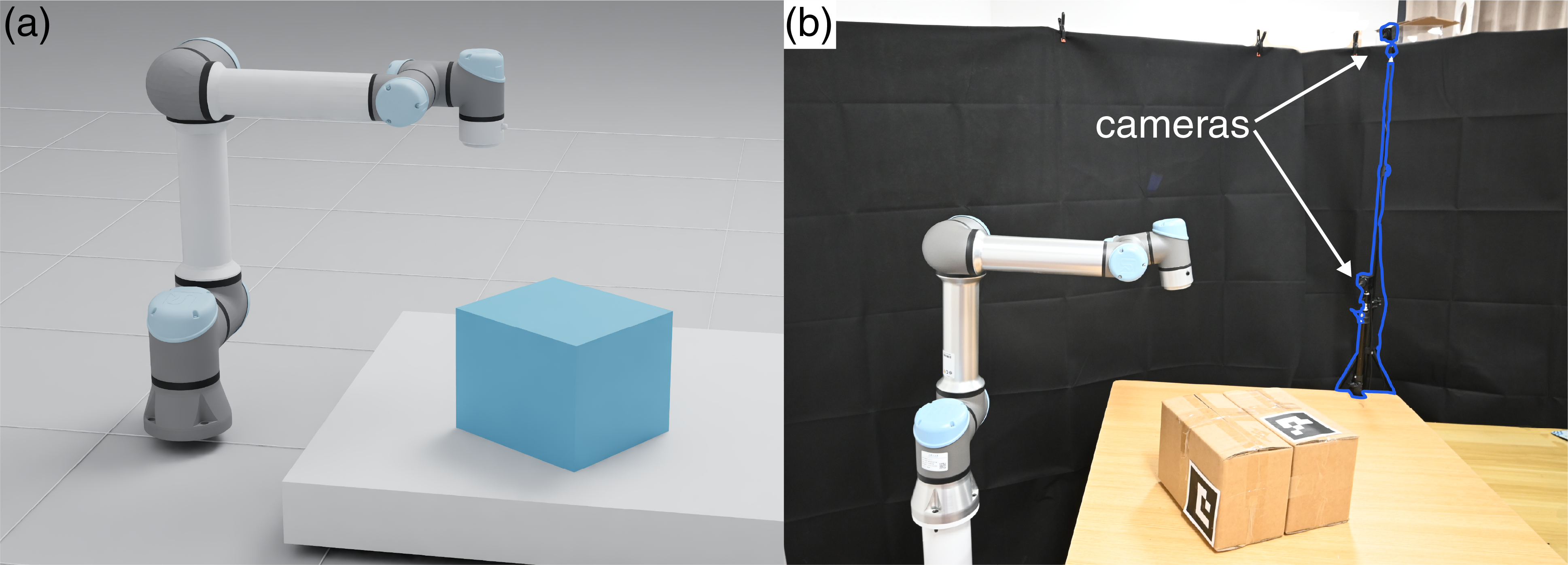}
    \caption{Experimental setups. (a) Isaac Lab simulation environment. (b) Physical robot setup, where two cameras detect ArUco markers attached to the box to track its position.}
    \label{fig:experimental_setup}
\end{figure}

The simulation and physical experiments use identical six axis Universal Robots UR7e robots. At each 30~Hz policy step, the six dimensional action specifies incremental position commands for the shoulder pan, shoulder lift, elbow, wrist~1, wrist~2, and wrist~3 joints. The resulting targets are tracked by a joint position PD controller. Simulation and current mapping run at 120~Hz, and each policy action is maintained for four simulation steps. Physical trials use two ordinary corrugated cardboard shipping boxes with dimensions $26\times18.5\times15.5$~cm and $31\times26\times18.5$~cm. Their measured dimensions are supplied to the policy as normalized task inputs. Policy checkpoints, normalization, object properties, camera calibration, and safety limits are fixed before testing. Simulation initial conditions and domain randomization use configured ranges and fixed seeds. For hardware evaluation, the box is placed at a new random pose before each trial. The corresponding simulation and hardware configurations are illustrated in Fig.~\ref{fig:experimental_setup}.

Complete task success is scored according to the criterion defined in Sec.~\ref{subsec:task_formulation}. In simulation, privileged state and contact signals additionally provide outcomes for each stage, failure categories, and unintended contact events. Hardware trials record whether the complete task criterion is satisfied, together with timeouts and safety events from controller logs; any timeout or safety event is counted as unsuccessful.

\subsection{Current Alignment Accuracy}
\label{subsec:exp_alignment}

The current alignment experiment evaluates offline alignment of physical motor current from synchronized simulation and robot data. The paired recording is split chronologically, with the first 80\% used for training and the final 20\% reserved for validation. A separate linear scaling from simulated joint effort to physical motor current is fitted for each joint. This baseline, an MLP, a recurrent GRU, and a Transformer are evaluated against the same held out current targets using MAE and RMSE for each joint and for all joints together.

\begin{figure}
    \centering
    \includegraphics[width=0.9\columnwidth]{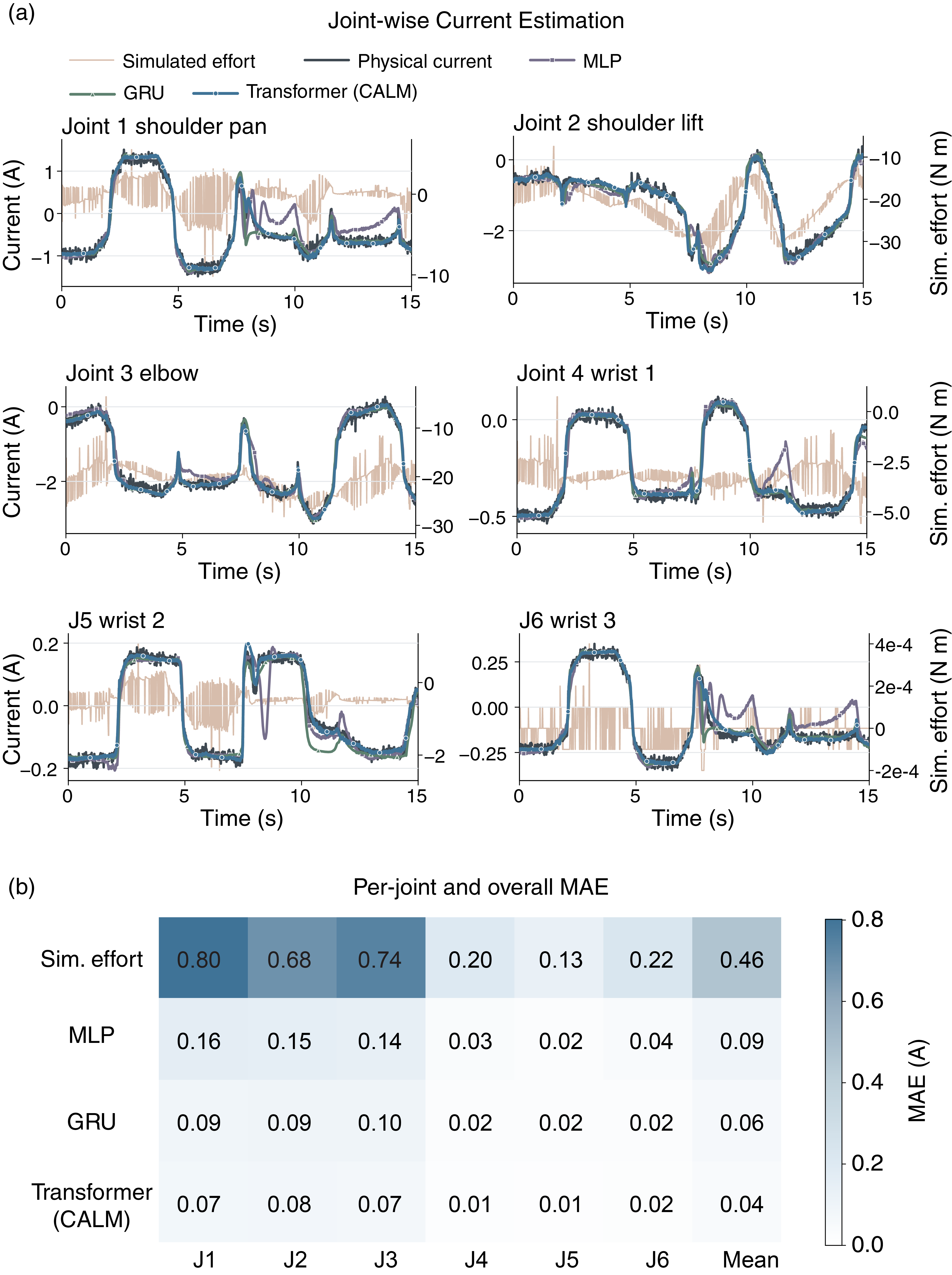}
    \caption{Offline current alignment. (a) Raw simulated joint effort (right axes), recorded physical motor current, and current predictions (left axes) over the first eligible continuous 15~s interval. (b) MAE for each joint and the mean across joints over all 19,213 held out samples. The simulated joint effort baseline is converted to amperes using a separate linear scaling for each joint fitted on the training data.}
    \label{fig:current_alignment_results}
\end{figure}

Across 19,213 held out samples from 65 contiguous segments of one recording session, the saved Transformer obtained an aggregate MAE of 0.04~A and RMSE of 0.07~A. The GRU was the strongest simulation only baseline, with an MAE of 0.06~A and RMSE of 0.12~A; the MLP obtained 0.09~A and 0.16~A, respectively, and the linearly scaled simulated joint effort obtained 0.46~A and 0.58~A. Transformer MAE ranged from 0.01~A for wrist~2 to 0.08~A for shoulder lift. These results are summarized in Fig.~\ref{fig:current_alignment_results}.

\subsection{Teacher Evaluation and Routed Rollouts}
\label{subsec:exp_teachers}

\begin{figure}[!t]
    \centering
    \includegraphics[width=0.85\columnwidth]{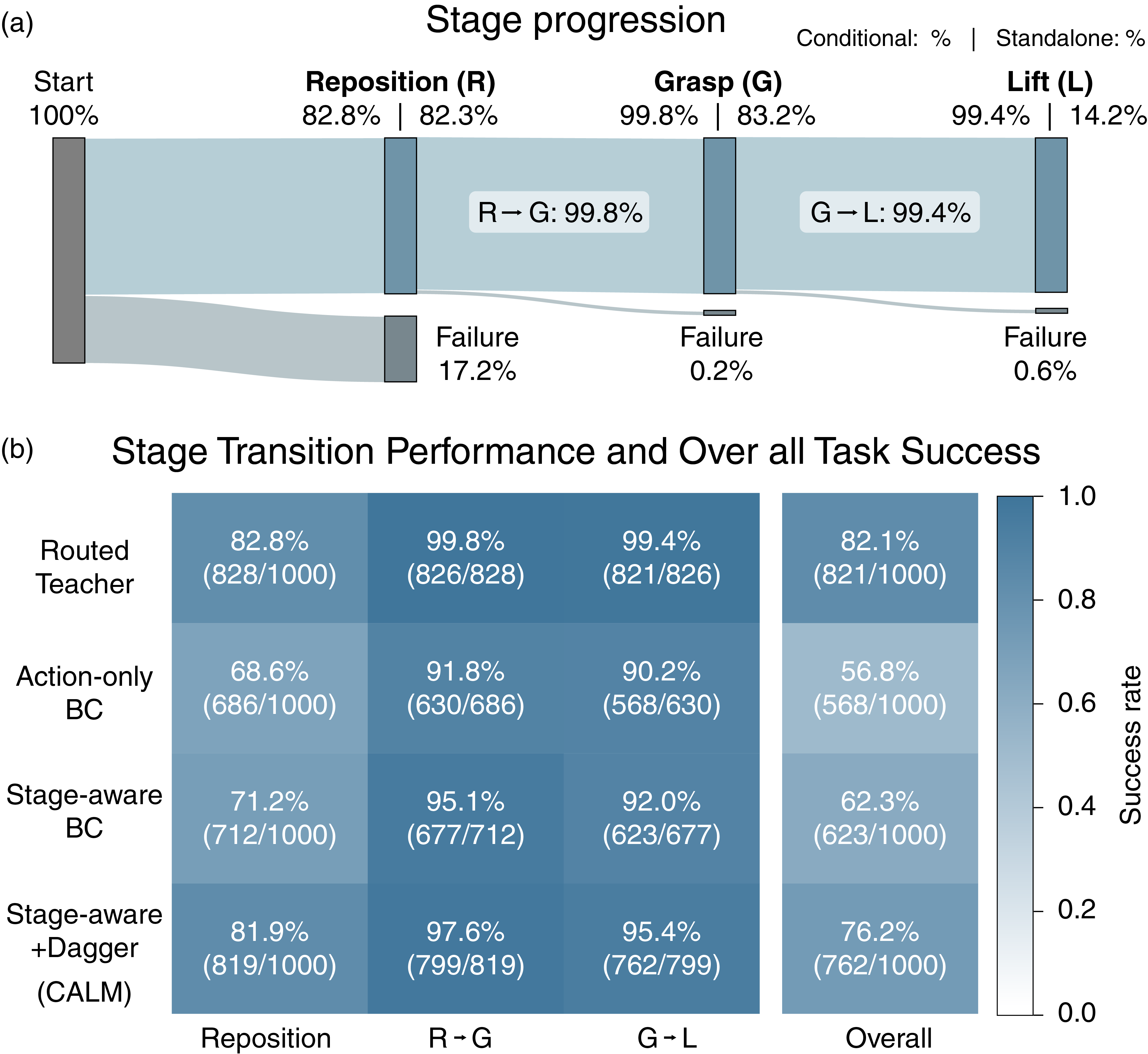}

    \caption{Teacher routing and teacher to student transfer in simulation. (a) Routed teacher stage progression. Each stage reports routed reach followed by independently evaluated teacher success, and each failure branch terminates at the first unreached stage. (b) Ablation of stage supervision and DAgger. Cells report unconditional reposition completion, conditional $R{\rightarrow}G$ and $G{\rightarrow}L$ success, and overall task success over 1,000 episodes per policy.}
    \label{fig:teacher_student_results}
\end{figure}

The frozen teachers are evaluated in two ways. (1) \emph{Independent evaluation}: each teacher is tested for 1,000 episodes using its stage specific reset states and terminal criterion. The reposition teacher starts from initial task states, the grasp teacher from stored successful reposition states, and the lift teacher from stored successful grasp states. (2) \emph{Routed evaluation}: the privileged router sequences the three teachers through one complete episode in each of 1,000 simulation environments. We report independent teacher success, unconditional routed stage reach, conditional handoff success, and routed completion. The results are shown in Fig.~\ref{fig:teacher_student_results}(a).

In the independent evaluations, the reposition, grasp, and lift teachers succeeded in 823/1,000, 832/1,000, and 142/1,000 episodes, respectively. Of the 1,000 routed episodes, 828 completed repositioning, 826 completed grasping, and 821 completed lifting, giving an overall routed completion rate of 82.1\%. Conditional completion was 826/828 (99.8\%) from repositioning to grasping and 821/826 (99.4\%) from grasping to lifting. All 179 unsuccessful routed episodes timed out: 172 before repositioning, two after repositioning but before grasping, and five after grasping. The low independent lifting success rate may partly result from its strict failure criterion: even a small initial slip of the box relative to the gripping links is classified as grasp loss, although the displacement may be difficult to detect visually. During routed evaluation, the router can return control to the grasp teacher to reestablish contact and then retry lifting, substantially improving complete task success.

\subsection{Teacher to Student Transfer}
\label{subsec:exp_student}

Figure~\ref{fig:teacher_student_results}(b) evaluates the contributions of stage supervision and DAgger through two ablations. First, action only BC is compared with stage aware BC to isolate the auxiliary stage objective. Second, stage aware BC is compared with its DAgger refined variant to evaluate correction using teacher labels from student visited states.

The routed teacher, action only BC, stage aware BC, and stage aware student refined with DAgger achieved complete task success rates of 82.1\%, 56.8\%, 62.3\%, and 76.2\%, respectively. Stage supervision increased success by 5.5 percentage points, while DAgger produced a further 13.9 point improvement and reduced the remaining gap to the routed teacher to 5.9 points. These run results support contributions from both stage supervision and DAgger refinement.

\subsection{Physical CALM Evaluation}
\label{subsec:exp_physical}

\begin{figure*}[!t]
    \centering
    \includegraphics[width=0.85\textwidth]{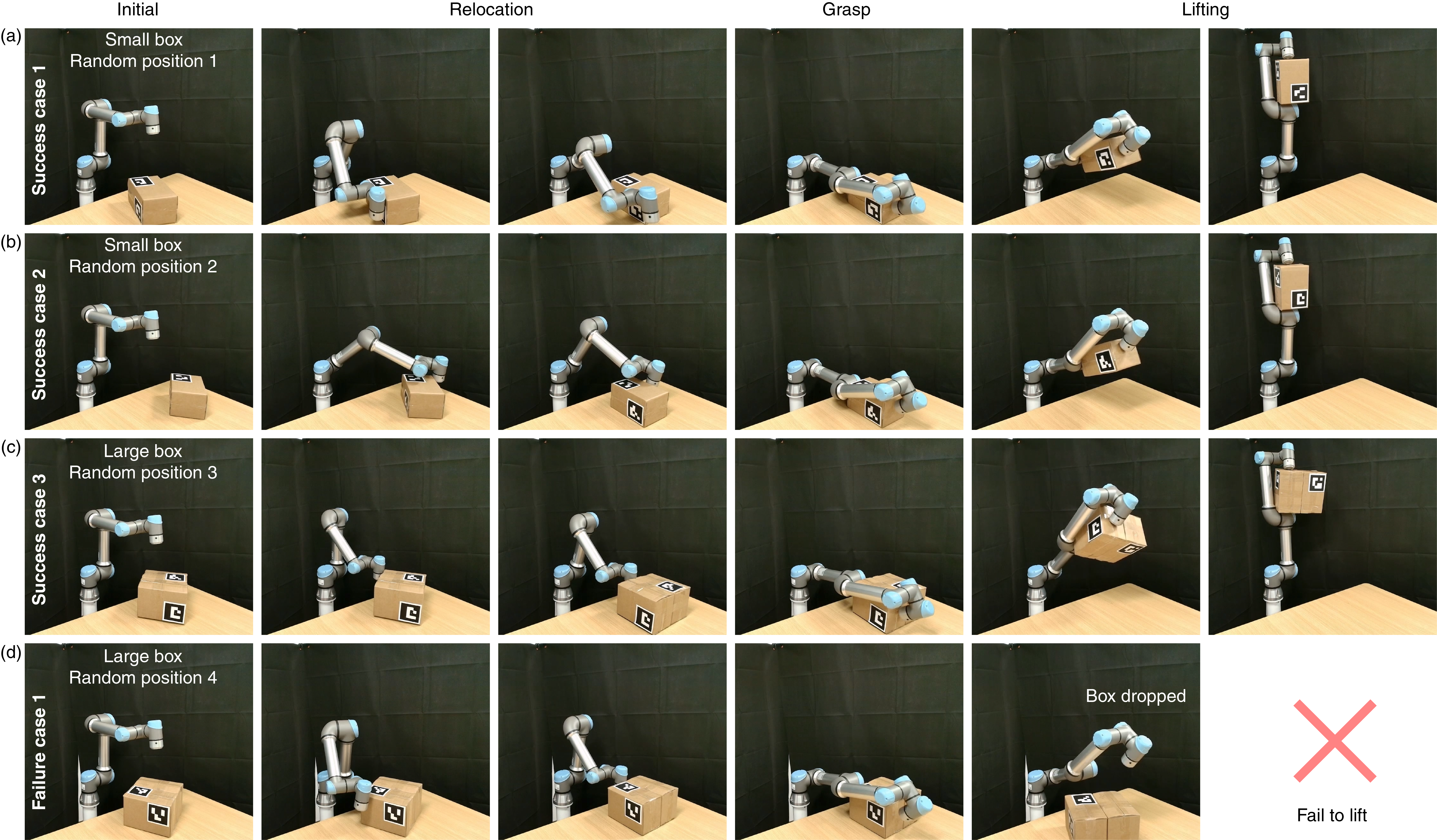}
    \caption{Physical CALM evaluation under random initial box placements. From left to right, each row shows the initial state, box repositioning, opposed link grasping, and the lifting outcome. (a) and (b) Successful trials with the small box. (c) Successful trial with the large box. (d) Failed trial in which the large box is dropped after grasping.}
    \label{fig:physical_experiment_sequence}
\end{figure*}

Physical experiments evaluate the complete task performance of the final CALM policy on hardware. To evaluate the contribution of current alignment, we train an ablated student using simulated joint-effort observations instead of mapper-predicted motor current. During hardware deployment, both students receive controller-measured motor currents. This ablation tests whether aligning the training observations with the physical current measurements improves sim-to-real transfer. The same hardware setup and random placement procedure are used for both policies. Representative task progress is shown in Fig.~\ref{fig:physical_experiment_sequence}.

CALM achieved 22/30 successful hardware trials (73.3\%), whereas the student without current alignment achieved 0/10. Each trial began with a new random box placement. For the ablated student, six trials triggered protective stops because of excessive force during grasping or lifting, two ended with the box being dropped, and two timed out during repositioning. These results support the benefit of current alignment for hardware transfer in the evaluated task.

\section{Conclusion and Outlook}
\label{sec:conclusion_outlook}

This work investigates how a rigid robot can use its single arm to lift an oversized object. To address the task complexity and the gap between simulation and reality, CALM trains three privileged teachers for repositioning, grasping, and lifting, then distills their routed demonstrations into a unified student using deployable observations. A Transformer based current mapper aligns simulated joint effort with physical motor current, providing the student with a consistent observation channel sensitive to loading in simulation and on hardware.

The present current mapper does not explicitly account for mobile base motion, or other coupled dynamics that alter motor current. Incorporating these factors could enable more versatile mobile manipulators for physically assistive work in human environments.

\bibliographystyle{IEEEtran}
\bibliography{references}

@article{bicchi1994,
  author  = {Antonio Bicchi},
  title   = {On the Problem of Decomposing Grasp and Manipulation Forces in Multiple Whole-Limb Manipulation},
  journal = {Robotics and Autonomous Systems},
  volume  = {13},
  number  = {2},
  pages   = {127--147},
  year    = {1994},
  doi     = {10.1016/0921-8890(94)90055-8}
}

@inproceedings{yuan2019,
  author    = {Weihao Yuan and Kaiyu Hang and Haoran Song and Danica Kragic and Michael Yu Wang and Johannes A. Stork},
  title     = {Reinforcement Learning in Topology-Based Representation for Human Body Movement with Whole Arm Manipulation},
  booktitle = {Proceedings of the IEEE International Conference on Robotics and Automation},
  pages     = {2153--2160},
  year      = {2019},
  doi       = {10.1109/ICRA.2019.8794160}
}

@inproceedings{florek2014,
  author    = {Monika Florek-Jasinska and Thomas Wimb{\"o}ck and Christian Ott},
  title     = {Humanoid Compliant Whole Arm Dexterous Manipulation: Control Design and Experiments},
  booktitle = {Proceedings of the IEEE/RSJ International Conference on Intelligent Robots and Systems},
  pages     = {1616--1621},
  year      = {2014},
  doi       = {10.1109/IROS.2014.6942771}
}

@article{seo2016,
  author  = {Jungwon Seo and Mark Yim and Vijay Kumar},
  title   = {A Theory on Grasping Objects Using Effectors with Curved Contact Surfaces and Its Application to Whole-Arm Grasping},
  journal = {The International Journal of Robotics Research},
  volume  = {35},
  number  = {9},
  pages   = {1080--1102},
  year    = {2016},
  doi     = {10.1177/0278364915600079}
}

@article{walker2013,
  author  = {Ian D. Walker},
  title   = {Continuous Backbone ``Continuum'' Robot Manipulators},
  journal = {ISRN Robotics},
  volume  = {2013},
  pages   = {726506},
  year    = {2013},
  doi     = {10.5402/2013/726506}
}

@inproceedings{punyo2022,
  author    = {Aimee Goncalves and Naveen Kuppuswamy and Andrew Beaulieu and Avinash Uttamchandani and Katherine M. Tsui and Alex Alspach},
  title     = {{Punyo-1}: Soft Tactile-Sensing Upper-Body Robot for Large Object Manipulation and Physical Human Interaction},
  booktitle = {Proceedings of the IEEE International Conference on Soft Robotics},
  pages     = {844--851},
  year      = {2022},
  doi       = {10.1109/RoboSoft54090.2022.9762117}
}

@article{barreiros2025,
  author  = {Jose A. Barreiros and Aykut {\"O}zg{\"u}n {\"O}nol and Mengchao Zhang and Sam Creasey and Aimee Goncalves and Andrew Beaulieu and Aditya Bhat and Katherine M. Tsui and Alex Alspach},
  title   = {Learning Contact-Rich Whole-Body Manipulation with Example-Guided Reinforcement Learning},
  journal = {Science Robotics},
  volume  = {10},
  number  = {105},
  pages   = {eads6790},
  year    = {2025},
  doi     = {10.1126/scirobotics.ads6790}
}

@article{jeon2024,
  author  = {Seunghun Jeon and Moonkyu Jung and Suyoung Choi and Beomjoon Kim and Jemin Hwangbo},
  title   = {Learning Whole-Body Manipulation for Quadrupedal Robot},
  journal = {IEEE Robotics and Automation Letters},
  volume  = {9},
  number  = {1},
  pages   = {699--706},
  year    = {2024},
  doi     = {10.1109/LRA.2023.3335777}
}

@inproceedings{robopanoptes2025,
  author    = {Xiaomeng Xu and Dominik Bauer and Shuran Song},
  title     = {{RoboPanoptes}: The All-Seeing Robot with Whole-Body Dexterity},
  booktitle = {Proceedings of Robotics: Science and Systems},
  year      = {2025},
  doi       = {10.15607/RSS.2025.XXI.042}
}

@inproceedings{uan2025,
  author    = {Nolan Fey and Gabriel B. Margolis and Martin Peticco and Pulkit Agrawal},
  title     = {Bridging the Sim-to-Real Gap for Athletic Loco-Manipulation},
  booktitle = {Proceedings of Robotics: Science and Systems},
  year      = {2025},
  doi       = {10.15607/RSS.2025.XXI.125}
}

@inproceedings{dagger2011,
  author    = {St{\'e}phane Ross and Geoffrey Gordon and Drew Bagnell},
  title     = {A Reduction of Imitation Learning and Structured Prediction to No-Regret Online Learning},
  booktitle = {Proceedings of the Fourteenth International Conference on Artificial Intelligence and Statistics},
  series    = {Proceedings of Machine Learning Research},
  volume    = {15},
  pages     = {627--635},
  year      = {2011}
}

@inproceedings{zhuang2023,
  author    = {Ziwen Zhuang and Zipeng Fu and Jianren Wang and Christopher G. Atkeson and S{\"o}ren Schwertfeger and Chelsea Finn and Hang Zhao},
  title     = {Robot Parkour Learning},
  booktitle = {Proceedings of the 7th Conference on Robot Learning},
  series    = {Proceedings of Machine Learning Research},
  volume    = {229},
  pages     = {73--92},
  year      = {2023}
}

@inproceedings{xu2025,
  author    = {Charles Xu and Qiyang Li and Jianlan Luo and Sergey Levine},
  title     = {{RLDG}: Robotic Generalist Policy Distillation via Reinforcement Learning},
  booktitle = {Proceedings of Robotics: Science and Systems},
  year      = {2025},
  doi       = {10.15607/RSS.2025.XXI.028}
}

@inproceedings{chen2023,
  author    = {Yuanpei Chen and Chen Wang and Fei-Fei Li and Karen Liu},
  title     = {Sequential Dexterity: Chaining Dexterous Policies for Long-Horizon Manipulation},
  booktitle = {Proceedings of the 7th Conference on Robot Learning},
  series    = {Proceedings of Machine Learning Research},
  volume    = {229},
  pages     = {3809--3829},
  year      = {2023}
}

@inproceedings{tan2018,
  author    = {Jie Tan and Tingnan Zhang and Erwin Coumans and Atil Iscen and Yunfei Bai and Danijar Hafner and Steven Bohez and Vincent Vanhoucke},
  title     = {Sim-to-Real: Learning Agile Locomotion for Quadruped Robots},
  booktitle = {Proceedings of Robotics: Science and Systems},
  year      = {2018},
  doi       = {10.15607/RSS.2018.XIV.010}
}

@article{hwangbo2019,
  author  = {Jemin Hwangbo and Joonho Lee and Alexey Dosovitskiy and Dario Bellicoso and Vassilios Tsounis and Vladlen Koltun and Marco Hutter},
  title   = {Learning Agile and Dynamic Motor Skills for Legged Robots},
  journal = {Science Robotics},
  volume  = {4},
  number  = {26},
  pages   = {eaau5872},
  year    = {2019},
  doi     = {10.1126/scirobotics.aau5872}
}

@article{wahrburg2018,
  author  = {Arne Wahrburg and Johannes Bos and Kim D. Listmann and Fan Dai and Bj{\"o}rn Matthias and Hao Ding},
  title   = {Motor-Current-Based Estimation of Cartesian Contact Forces and Torques for Robotic Manipulators and Its Application to Force Control},
  journal = {IEEE Transactions on Automation Science and Engineering},
  volume  = {15},
  number  = {2},
  pages   = {879--886},
  year    = {2018},
  doi     = {10.1109/TASE.2017.2691136}
}

@article{callar2023,
  author  = {Tolga-Can {\c{C}}allar and Sven B{\"o}ttger},
  title   = {Hybrid Learning of Time-Series Inverse Dynamics Models for Locally Isotropic Robot Motion},
  journal = {IEEE Robotics and Automation Letters},
  volume  = {8},
  number  = {2},
  pages   = {1061--1068},
  year    = {2023},
  doi     = {10.1109/LRA.2022.3222951}
}

@inproceedings{mohammad2023,
  author    = {Aran Mohammad and Moritz Schappler and Tobias Ortmaier},
  title     = {Towards Human-Robot Collaboration with Parallel Robots by Kinetostatic Analysis, Impedance Control and Contact Detection},
  booktitle = {Proceedings of the IEEE International Conference on Robotics and Automation},
  pages     = {12092--12098},
  year      = {2023},
  doi       = {10.1109/ICRA48891.2023.10161217}
}

@article{argall2009,
  author  = {Brenna D. Argall and Sonia Chernova and Manuela Veloso and Brett Browning},
  title   = {A Survey of Robot Learning from Demonstration},
  journal = {Robotics and Autonomous Systems},
  volume  = {57},
  number  = {5},
  pages   = {469--483},
  year    = {2009},
  doi     = {10.1016/j.robot.2008.10.024}
}

@article{kober2013,
  author  = {Jens Kober and J. Andrew Bagnell and Jan Peters},
  title   = {Reinforcement Learning in Robotics: A Survey},
  journal = {The International Journal of Robotics Research},
  volume  = {32},
  number  = {11},
  pages   = {1238--1274},
  year    = {2013},
  doi     = {10.1177/0278364913495721}
}

@inproceedings{yu2025mimictouch,
  author    = {Kelin Yu and Yunhai Han and Qixian Wang and Vaibhav Saxena and Danfei Xu and Ye Zhao},
  title     = {{MimicTouch}: Leveraging Multi-modal Human Tactile Demonstrations for Contact-rich Manipulation},
  booktitle = {Proceedings of the 8th Conference on Robot Learning},
  series    = {Proceedings of Machine Learning Research},
  volume    = {270},
  pages     = {4844--4865},
  year      = {2025}
}

@article{glick2018,
  author  = {Paul E. Glick and Srinivasan A. Suresh and Donald Ruffatto and Mark R. Cutkosky and Michael T. Tolley and Aaron Parness},
  title   = {A Soft Robotic Gripper With Gecko-Inspired Adhesive},
  journal = {IEEE Robotics and Automation Letters},
  volume  = {3},
  number  = {2},
  pages   = {903--910},
  year    = {2018},
  doi     = {10.1109/LRA.2018.2792688}
}

@article{zhou2026,
  author  = {Meijie Zhou and Jianjun Yuan and Zhengtao Hu and Sheng Bao and Liang Du},
  title   = {{1D2L}: One-Arm Drag and Two-Arm Lift for Manipulating Large and Heavy Tabletop Objects},
  journal = {IEEE Robotics and Automation Letters},
  volume  = {11},
  number  = {6},
  pages   = {7262--7269},
  year    = {2026},
  doi     = {10.1109/LRA.2026.3685936}
}

@article{zhou2026vision,
  author        = {Peng Zhou and Jun Hu and Sihan Chen and Zeqing Zhang and Haofei Ma and Zhenyu Lu and Sichao Liu and Xueqian Wang and Pai Zheng and Xiang Li and Shan Luo and Jia Pan and David Navarro-Alarcon and Chenguang Yang and Michael Yu Wang},
  title         = {Vision-Based Tactile Intelligence for Robotics: Sensing, Learning, and Embodied Manipulation},
  journal       = {arXiv preprint arXiv:2608.15490},
  year          = {2026},
  eprint        = {2608.15490},
  archivePrefix = {arXiv},
  primaryClass  = {cs.RO}
}

@inproceedings{vaswani2017attention,
  author    = {Ashish Vaswani and Noam Shazeer and Niki Parmar and Jakob Uszkoreit and Llion Jones and Aidan N. Gomez and Lukasz Kaiser and Illia Polosukhin},
  title     = {Attention Is All You Need},
  booktitle = {Advances in Neural Information Processing Systems},
  volume    = {30},
  pages     = {5998--6008},
  year      = {2017}
}

\end{document}